\documentclass[10pt,twocolumn,letterpaper]{article}

\usepackage{iccv}
\usepackage{times}
\usepackage{epsfig}
\usepackage{graphicx}
\usepackage{amsmath}
\usepackage{amssymb}
\usepackage{float}
\usepackage{array}
\usepackage{bbm}
\usepackage{multirow}
\usepackage{pifont}
\usepackage{colortbl}
\usepackage{booktabs}
\usepackage{enumerate}
\usepackage{url}
\usepackage[T1]{fontenc}
\usepackage[utf8]{inputenc}
\usepackage{authblk}
\usepackage{stfloats}
\usepackage{fancyhdr}

\newcommand{\minus}{\scalebox{0.75}[1.0]{$-$}}
\newcommand{\cmark}{\ding{51}}%

\newcommand*\samethanks[1][\value{footnote}]{\footnotemark[#1]}
\makeatletter
\renewcommand\AB@affilsepx{\protect\Affilfont}

\makeatother

\usepackage[pagebackref=true,breaklinks=true,letterpaper=true,colorlinks,bookmarks=false]{hyperref}

\def\iccvPaperID{2244}

\iccvfinalcopy

\ificcvfinal\fi

\begin{document}

\title{Mining Latent Classes for Few-shot Segmentation\vspace{-0.5cm}}

\author{Lihe Yang\textsuperscript{\rm 1}~~~ Wei Zhuo\textsuperscript{\rm 2}\thanks{Corresponding author.}~~~ Lei Qi\textsuperscript{\rm 3,1}~~~ Yinghuan Shi\textsuperscript{\rm 1}\samethanks[1]\hspace{0.15cm}\thanks{The work of Yinghuan Shi and Lihe Yang was supported by National Key Research and Development Program of China (2019YFC0118300). The work of Lei Qi was supported by China Postdoctoral Science Foundation funded project (2021M690609).}~~~ Yang Gao\textsuperscript{\rm 1}\\
\vspace{-0.3cm}

\fontsize{11pt}{\baselineskip}\selectfont{\textsuperscript{\rm 1}State Key Laboratory for Novel Software Technology, Nanjing University\\
\textsuperscript{\rm 2}Tencent\\
\textsuperscript{\rm 3}Key Lab of Computer Network and Information Integration (Ministry of Education), Southeast University\\}
\vspace{0.1cm}

\small \texttt{lihe.yang.cs@gmail.com~~ weizhuo@tencent.com~~ qilei@seu.edu.cn~~ \{syh, gaoy\}@nju.edu.cn}}

\maketitle
\ificcvfinal\thispagestyle{empty}\fi

\thispagestyle{fancy}
\lhead{}
\chead{}
\rhead{}
\lfoot{}
\rfoot{}
\cfoot{\thepage}
\renewcommand{\headrulewidth}{0pt}
\renewcommand{\footrulewidth}{0pt}
\pagestyle{fancy}
\cfoot{\thepage}

\begin{abstract}
    Few-shot segmentation (FSS) aims to segment unseen classes given only a few annotated samples. Existing methods suffer the problem of feature undermining, i.e., potential novel classes are treated as background during training phase. Our method aims to alleviate this problem and enhance the feature embedding on latent novel classes. In our work, we propose a novel joint-training framework. Based on conventional episodic training on support-query pairs, we introduce an additional mining branch that exploits latent novel classes via transferable sub-clusters, and a new rectification technique on both background and foreground categories to enforce more stable prototypes. Over and above that, our transferable sub-cluster has the ability to leverage extra unlabeled data for further feature enhancement. Extensive experiments on two FSS benchmarks demonstrate that our method outperforms previous state-of-the-art by a large margin of 3.7\% mIOU on PASCAL-5$^i$ and 7.0\% mIOU on COCO-20$^i$ at the cost of 74\% fewer parameters and 2.5x faster inference speed. The source code is available at \url{https://github.com/LiheYoung/MiningFSS}.
\end{abstract}
\section{Introduction}

Advanced by fully convolutional neural networks, semantic segmentation has achieved impressive progress \cite{long2015fully, zhao2017pyramid, chen2018encoder, li2019expectation, yuan2019object}. Nevertheless, fully-supervised semantic segmentation demands a large amount of pixel-wise annotations which are exhaustive to acquire. This problem urges the need for few-shot segmentation where only a handful of annotations are required for novel classes. In this setting, however, methods with conventional training paradigm \cite{long2015fully} easily suffer overfitting. In view of this, recent FSS works aim to learn a generic manner from seen classes and adapt to the unseen classes via the few shots, namely supports.

\begin{figure}
    \centering
    \includegraphics[width=0.45\textwidth]{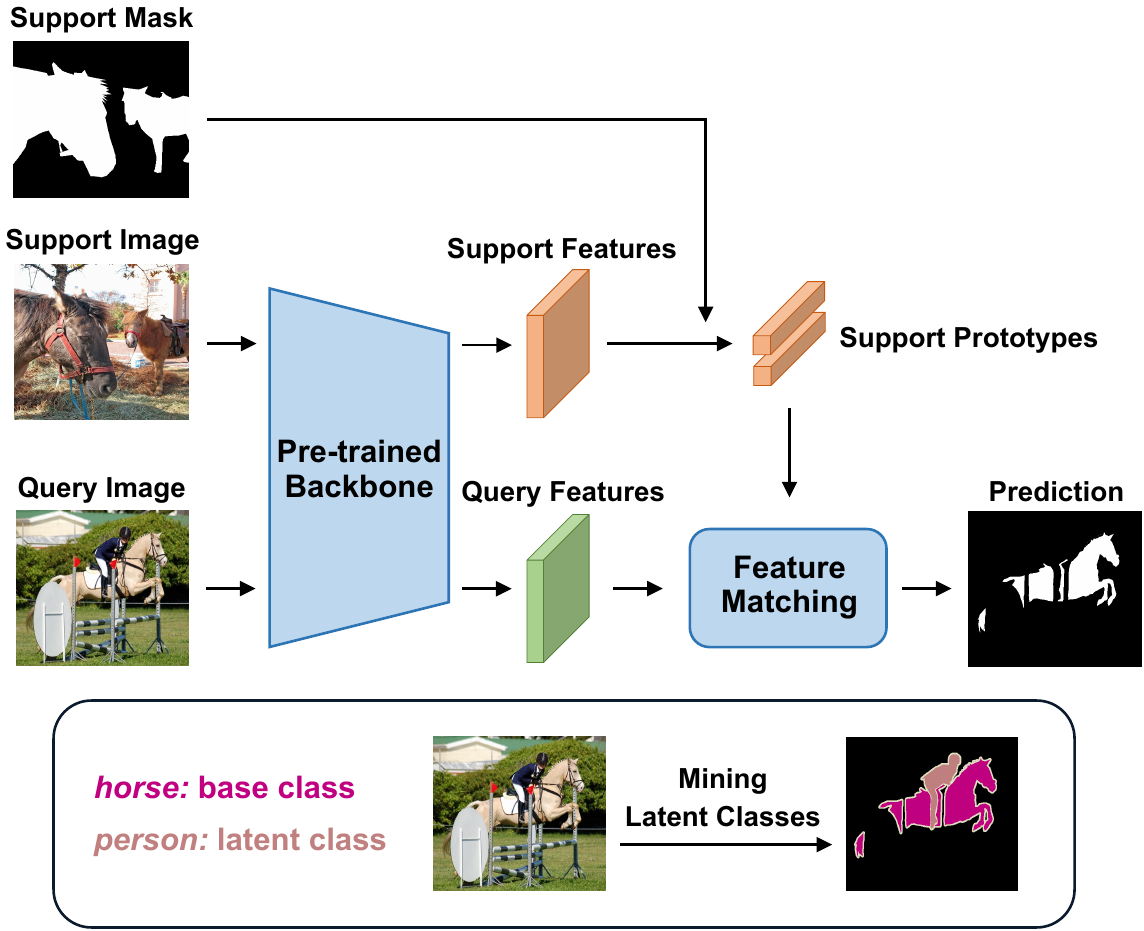}
    \caption{Illustration of the few-shot segmentation framework and the latent novel classes in the background. Typical FSS setting is only concerned about the current support class, and treats other latent classes as background in each episode of the training. The latent classes, however, are fundamentally different from the real backgrounds, and deserve better exploitation.
    }
    \label{intro_fig}
    \vspace{-0.3cm}
\end{figure}

The recent research on few-shot segmentation \cite{shaban2017one, zhang2018sg, wang2019panet, liu2020part, tian2020prior} has gained some progress. Shaban \etal \cite{shaban2017one} first proposed a siamese network on support-query pairs and alleviated the overfitting problem. Later works developed non-parametric bidirectional alignment mechanisms \cite{wang2019panet}, fine-grained part-aware prototypes \cite{liu2020part}, multi-scale feature enhancement modules \cite{tian2020prior} \etc. 

Despite their success, we notice that these methods rarely exploit the inherent problems of FSS, namely the \textit{feature undermining} problem and the \textit{prototype bias} problem: 1) the feature undermining problem is that embeddings of the latent novel classes are over-smoothed when learnt as the background in typical FSS. As shown in Figure \ref{intro_fig}, only current support class is concerned about in each episode, the latent novel class \textsl{person} is incorrectly treated as background. 2) The prototype bias problem is caused by the fact that few shots cannot mimic the real class-wise statistics, making it sub-optimal to merely utilize the current supports for prototype estimation. In our work, we aim to exploit the latent novel classes and develop prototypes with less bias to narrow down the gap between few shots and real statistics. 

For the exploitation of latent novel class, a related field is the self-supervised learning, that defines pretexts such as solving jigsaws \cite{noroozi2016unsupervised}, predicting rotations \cite{gidaris2018unsupervised} and discriminating instances \cite{he2020momentum, chen2020simple}, to mine the unlabeled open set images. These methods, mainly work as pre-trained models and still require sufficient data for other downstream tasks, such as detection, segmentation \etc. This is mainly due to that more features on finer scales are required which are not aimed at by current self-supervised techniques. Beyond self-supervised learning, semi-supervised learning also exploits the unlabeled data that has the same category scope with the labeled data. It, however, is not aimed at knowledge transfer, and it cannot mine latent classes explicitly which are disjoint with known classes.

For the prototype bias problem, PGNet \cite{zhang2019pyramid} develops an attention module based on pyramid graphs to fuse support features. PPNet \cite{liu2020part} attempts to modify support prototypes based on superpixels from extra samples. However, they do not take full use of whole training set. In addition, the prototype bias, \ie background features from the supports, is rarely tackled exclusively from its inherent characteristics.

In our work, we propose a novel latent class mining strategy with pseudo labeling, and a novel prototype rectification technique, based on the metric learning framework on support-query pairwise inputs. We consider \textit{every pixel matters} in the training set, which implies that even the temporally annotated backgrounds can contain novel classes, and an explicit mining can enhance the feature discrimination. In particular, 1) our auxiliary branch exploits latent novel classes from the backgrounds in the training set via semantic sub-clusters transferred from the annotated base classes. More than that, our method can leverage extra unlabeled data for further feature enhancement. Note that, our method is also well fit for more realistic settings, where plenty of additional novel classes may exist due to limited labor for annotation or the fact that novel classes have not been required or discovered while labeling. 2) On the other hand, we propose a novel technique to rectify the prototypes of both the foreground classes and the background. As aforementioned, we suppose background takes much more information than the support prior, and we propose to model the background via broader set, namely the whole training set, via an moving average. In addition, we improve PPNet \cite{liu2020part} by incorporating more stable region features for the foreground prototype rectification. 

In summary, our contributions lie in four folds:
\begin{itemize}
\setlength{\itemsep}{0pt}
\setlength{\parsep}{0pt}
\setlength{\parskip}{0pt}
\item We propose a novel framework that mines latent object and learns the pairwise metric jointly. Taking advantage of the novel framework, our method can be applied to unseen classes directly without further training or fine-tuning, and meanwhile it does not suffer the feature undermining problem. 

\item We propose a novel prototype rectification technique to alleviate the prototype bias problem by incorporating a stable global background prototype and relevant foreground region neighbors. 

\item We conduct extensive experiments proving that our model takes fewer parameters, evaluates at faster speed, and achieves better performance. 

\item Extension experiments on the unlabeled data from different sources prove that our latent class mining method can exploit unlabeled data and boost the performance further.

\end{itemize}

\section{Related Work}

\textbf{Semantic Segmentation.} 
Semantic segmentation that provides pixel-wise dense semantic prediction has gained interests in computer vision community for decades \cite{shi2000normalized, comaniciu2002mean, ren2012rgb}. Inspired by the success of fully convolutional networks \cite{long2015fully} that train an end-to-end network for segmentation, later works \cite{zhao2017pyramid, chen2018encoder, zhang2018context, huang2019ccnet, yuan2019object,chen2018encoder, ronneberger2015u, lin2017refinenet, sun2019high} contribute many benchmark blocks, such as the pyramid pooling module \cite{zhao2017pyramid}, dilated convolution \cite{chen2018encoder}, deformable convolution \cite{dai2017deformable}, non-local module \cite{wang2018non, zhu2019asymmetric} \etc. Thanks to these blocks, current semantic segmentation performance has been greatly improved.
The traditional scenario, however, usually requires plenty of data which is costly. In our work, we focus on the semantic segmentation in few-shot scenario. 

\textbf{Few-shot Learning.} 
Few-shot learning (FSL), due to its low cost for application, has gained interests for many years. Recognizing unseen classes with few shots is meaningful, but also challenging. To this end, a stream of works in meta learning \cite{finn2017model,snell2017prototypical, ren2018meta} are proposed to extract meta knowledge that are assumed to be shared among the known and unseen classes. A majority of recent works follow this research line, and these works can be further divided into three folds that are the model-based methods \cite{santoro2016one, munkhdalai2017meta}, the optimization-based methods \cite{finn2017model, nichol2018first, ravi2016optimization} and the metric-based methods \cite{vinyals2016matching, snell2017prototypical, sung2018learning}. Even though the few-shot learning, mainly on classification, has been extensively exploited, it cannot be easily adapted to segmentation due to the dense prediction problem. It is worth noting that Liu \etal \cite{liu2019prototype} rectifies the support prototypes in FSL, but our rectification technique is also especially designed for the background class, which is unique in segmentation task.

\textbf{Few-shot Segmentation.} 
The few-shot segmentation \cite{shaban2017one, zhang2018sg, wang2019panet, siam2019amp, tian2020differentiable, liu2020part, liu2020crnet, tian2020prior, boudiaf2020few, ouyang2020self, li2020fss, azad2021texture} has received considerable attention very recently. Inspired by the few-shot learning, Shaban \etal \cite{shaban2017one} contributes the first few-shot segmentation work, whose segmentation parameters are generated by a conditioning branch on the supports. Different from \cite{shaban2017one}, a later work \cite{zhang2018sg} generates the foreground object segmentation of the support class by measuring the embedding similarity between query and supports, where their embeddings are extracted by the same backbone model. PANet \cite{wang2019panet} extends this work to align the support and query bidirectionally where each can be the reference for the other. Compared with the above works that only use a holistic prototype for each category in supports, PPNet \cite{liu2020part} adopts part-aware prototypes to capture the diverse fine-grained object features. As mentioned before,  existing methods merely treat the classes not belonging to base classes as the background and suffer the problem of feature undermining. Motivated by this, we boost the few-shot segmentation via mining latent objects from the backgrounds.

\textbf{Semi-/self-supervised Learning.}
In term of pseudo labeling and leveraging the unlabeled data, we will briefly review the semi-/self-supervised methods here. The semi-supervised methods include consistency regularization \cite{tarvainen2017mean, berthelot2019mixmatch, sohn2020fixmatch, ouali2020semi}, entropy minimization \cite{grandvalet2005semi, saito2019semi}, pseudo labeling \cite{lee2013pseudo, Li_2020_CVPR} \etc. However, conventional pseudo labeling strategy works under the hypothesis that the unlabeled are of the same class space as the labeled. In another research line, the self-supervised learning attempts to learn purely on unlabeled data \cite{noroozi2016unsupervised, gidaris2018unsupervised} or serve as an auxiliary supervision on training data \cite{gidaris2019boosting, zhai2019s4l}.
Recently, contrastive learning based methods \cite{he2020momentum,chen2020simple} even perform on-par with the supervised counterparts in classification. They enforce the variations of any crops in an image to be consistent, which is however contradictory to the target of segmentation that requires discriminative features on regions. Different from self-supervised learning, our method enforces multi-scale, \ie pixel-level and region-level supervision. 
\section{Method}
\label{method}

\subsection{Problem Definition}

The aim of few-shot segmentation is to obtain a model from base classes and the model can segment an unseen semantic class without re-training based on only a handful of labeled images of the unseen class.  Typically, in few-shot segmentation, a training set $\mathcal{D}_{tr}$ and a testing set $\mathcal{D}_{te}$ are given from two disjoint class sets $\mathcal{C}_{tr}$ and $\mathcal{C}_{te}$ individually. In particular, 
$\mathcal{D}_{tr} = \{(I_i, M_i)\}_{i=1}^{N_{tr}}$
is composed of $N_{tr}$ image-mask pairs that contain objects from $\mathcal{C}_{tr}$, where $I_i$ indicates the $i$-th image and $M_i$ is its corresponding mask. The testing set $\mathcal{D}_{te}$ is constructed in a similar way except that its targets are from classes $\mathcal{C}_{te}$.
A general application of few-shot segmentation works as that it collects a small \textit{support set} $\mathcal{S} = \{(I_i^s, M_i^s)\}_{i=1}^K$ with $K$ image-mask pairs of category $c$, and uses them to segment the objects of that category in the \textit{query set} $\mathcal{Q}$. To imitate the application process during training, a set of \textit{episodes} $\mathcal{E} = \{(\mathcal{S}_i, \mathcal{Q}_i)\}_{i=1}^{N_e}$ are randomly sampled from $\mathcal{D}_{tr}$. In each episode, the model makes prediction on query set $\mathcal{Q}_i$ conditioned on the support set $\mathcal{S}_i$. Here, $\mathcal{Q}_i = \{(I^q, M^q)\}$ is provided with ground-truth mask to supervise the training process. 

\subsection{Overview}
As aforementioned, the latent novel classes, not belonging to the pre-defined base classes, are simply learnt as the background during training, making existing methods sub-optimal in leveraging the training data. Motivated by this observation, we propose to mine the latent novel classes from the backgrounds to enhance the feature embeddings for better generalization to novel classes. Above that, we introduce a novel rectification technique for more stable and informative prototypes.

\textbf{Our Framework.}
We build a unified framework that conducts the meta learning via episodic training on support-query pairs, and meantime mines the latent novel classes from the backgrounds via an auxiliary supervision. With this joint training framework, our method can learn both transferable meta knowledge and promising embedding.

\begin{figure*}
\centering
\includegraphics[width=0.95\textwidth]{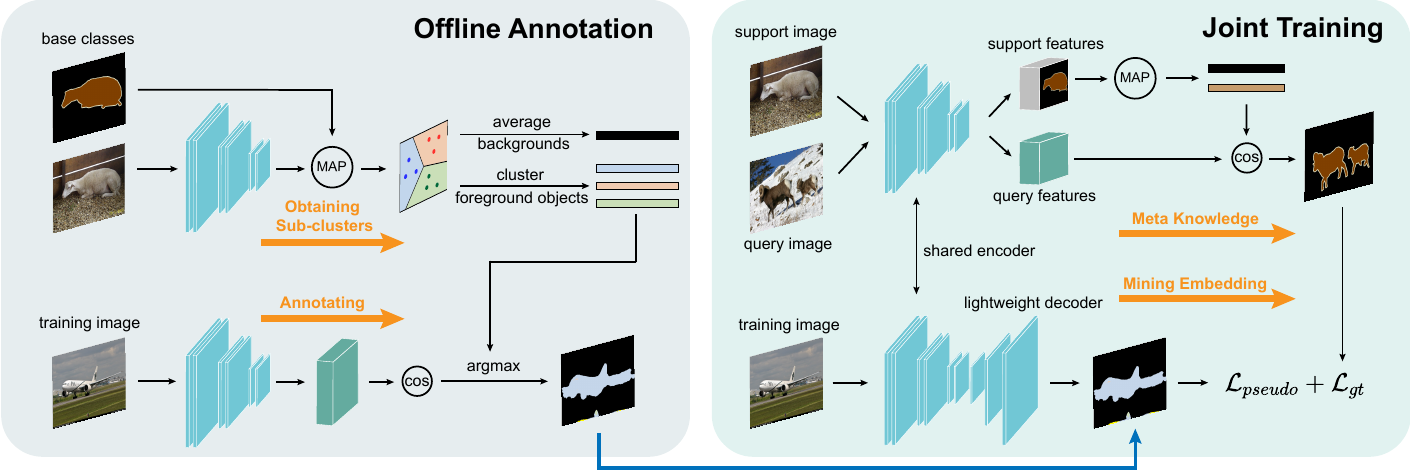}
\caption{The overall training pipeline of our method. The left part illustrates the offline annotation process while the right one illustrates the joint training process. Representative sub-clusters are produced via clustering prototypes of foreground objects from base classes and averaging all background prototypes (top left). With these semantic sub-clusters, we annotate training images densely by a nearest neighbor mapping strategy (bottom left). Given the pseudo masks of training images, the model is jointly optimized by support-query pairs with their groundtruth masks as well as extra sampled images with their pseudo masks (right).}
\label{network}
\vspace{-0.2cm}
\end{figure*}

To obtain the auxiliary supervision for latent classes, the training images are annotated with the \textit{representative sub-clusters} transferred from annotated base classes. The offline annotating process is only conducted once and the pseudo masks are kept the same during the whole training phase. A pipeline of our training process is shown in Figure \ref{network}. 

In episodic training, two kinds of inputs, \ie supports and a query, are first forwarded to a siamese network for feature extraction. Then, each query feature is compared with the prototype of current support class and the background prototype for classification. The prototypes are generated in a non-parametric manner of mask average pooling (MAP) on the extracted features. Here the segmentation is for binary classification of a support class or not. In our work, an additional supervision from pseudo labels of extra sampled images from the training set is added for multi-class segmentation. The overall optimization target can be briefly formulated as:
\vspace{-0.1cm}
\begin{equation}
\vspace{-0.1cm}
    \mathcal{L} = \mathcal{L}_{gt} + \lambda\mathcal{L}_{pseudo},
\label{loss}
\end{equation}
where $\lambda$ is the balance weight and simply set as $1$. 

Moreover, for a stable and informative estimation of support prototypes, we rectify background and foreground prototypes respectively via taking full advantage of the statistics in the training set. Specifically, the background prototype is rectified with a global one, which is maintained and updated during training to capture the common characteristics of various scenes, while the foreground prototype is rectified with the most relevant regions in the training set only during inference.

\subsection{Mining and Learning Latent Classes}
It can be summarized as a two-stage process, namely 1) \textit{pseudo labeling latent classes} and 2) \textit{learning latent classes}. It is feasible to annotate latent classes via the representative sub-clusters transferred from the base classes based on the assumption that foreground objects from the same domain should share some commonalities more or less with each other. For example, \textit{horse} and \textit{cow} might share commonality on shapes since they are both \textit {four leg's animals}. We acquire these transferable commonalities by grouping foreground objects and generate the semantic sub-clusters. The training images are then supervised jointly with both original groundtruth masks for transferable meta knowledge and the generated pseudo masks for discriminative embeddings on the latent novel classes.

\vspace{-0.15cm}
\subsubsection{Annotating with Representative Sub-clusters}

\textbf{Extracting representative sub-clusters.}  Given a pre-trained embedding network, we adopt the masked average pooling (MAP) strategy \cite{siam2019amp} to obtain a holistic description of a specific category in an image. The prototype 
$\mathbf{p}^c_i \in \mathbb{R}^C$ 
of the $c$-th class in the $i$-th image is computed as:
\vspace{-0.1cm}
\begin{equation}
\vspace{-0.1cm}
    \mathbf{p}^c_i = \frac{\sum_{x, y}F_i^{x, y}\mathbbm{1}[M_i^{x, y}=c ]}{\sum_{x, y}\mathbbm{1}[M_i^{x, y}=c ]},
\end{equation}
where $F_i \in \mathbb{R}^{C\times H\times W}$ is the extracted feature of the $i$-th image. Note that, $c$ ranges from $0$ to $|\mathcal{C}_{tr}|$ and $c = 0$ indicates the background class.

$\mathcal{P}_{fg}$ and $\mathcal{P}_{bg}$ denote the set of foreground and background prototypes in all annotated training images respectively. The K-Means clustering algorithm is performed on $\mathcal{P}_{fg}$ to produce $K$ most representative sub-clusters (cluster centers) $\mathcal{P}_{cluster}$, which are expected to capture some commonalities among various foreground objects. On the other hand, considering the backgrounds vary greatly from image to image, we simply average all prototypes in $\mathcal{P}_{bg}$ to produce single background prototype $\mathbf{p}_{bg}$, serving as a global descriptor of the backgrounds. Finally, a union set of $K+1$ representative sub-clusters $\mathcal{P}_{rep}$ are obtained by combining $\mathcal{P}_{cluster}$ and $\{\mathbf{p}_{bg}\}$, which can be viewed as high-level descriptors of backgrounds and foreground classes in $\mathcal{D}_{tr}$.

\vspace{0.2cm}
\noindent
\textbf{Annotating training images.} Here, we describe how to annotate training images according to $\mathcal{P}_{rep}$. For a training image $I \in \mathbb{R}^{3\times H\times W}$, its features extracted from the encoding network are denoted as $F \in \mathbb{R}^{C\times H\times W}$. The pseudo mask $M^p \in \mathbb{R}^{H\times W}$ can be obtained by performing dense classification on $F$. Next, we demonstrate why and how to classify each pixel based on nearest neighbor. 

We assume that, objects in $I$ share some commonalities with certain foreground prototype in $\mathcal{P}_{rep}$ while non-object area in $I$ may be closer to the background prototype in $\mathcal{P}_{rep}$.
Therefore, we measure the similarity between each feature $F(x,y)$ in $F$ and the $K+1$ representative prototypes in $\mathcal{P}_{rep}$, and classify $F(x, y)$ into one of $K+1$ categories, which can be formulated as:
\vspace{-0.1cm}
\begin{equation}
\vspace{-0.1cm}
    M^p(x, y) = \mathop{\arg\max}_{k}\ {\rm cos}\big(F(x, y), \mathbf{p}_k\big),
\end{equation}
where $\mathbf{p}_k \in \mathcal{P}_{rep}$ and ${\rm cos}(\cdot, \cdot)$ measures the cosine similarity between two vectors.

The labels in the obtained pseudo mask $M^p$ contain at most $K+1$ categories. It is worthy noting that except the background class, other $K$ clustered classes do not stand for any concrete objects or classes but they may represent several typical characteristics of actual existing categories. The produced pseudo masks segment the whole scene into several regions which contain inherent semantic consistency and can be utilized to learn more discriminative features. 

\subsubsection{Joint Training}
Given training images annotated with pseudo masks as well as ground-truth masks, we trained the encoding network with these two sources of supervision together. A mini-batch is constructed of both images with ground-truth masks and extra sampled images with pseudo masks.

For images with groundtruth masks, episodic training paradigm for meta learning is adopted to learn meta knowledge for quick adaptation to novel classes. In our model, we adopt a non-parametric matching mechanism similar to \cite{wang2019panet, liu2020part}. Cosine similarity function is applied to measure the similarity between each feature in the query image and the foreground and background prototypes from the support set. Loss function on a query image can be formulated as:
\vspace{-0.15cm}
\begin{equation}
\vspace{-0.15cm}
    \mathcal{L}_{gt} = \frac{1}{HW}\sum_{x, y}\sum_{c}\mathbbm{1}[M^{x, y} = c]\log \hat{M}_c^{x, y},
\end{equation}
and the score map $\hat{M}_c^{x, y}$ is defined by:
\vspace{-0.15cm}
\begin{equation}
\vspace{-0.15cm}
    \hat{M}_c^{x, y} = \frac { \exp\big({\rm cos}(F^{x, y}, \mathbf{p}^c) \cdot \sigma\big) } {\sum_q \exp\big({\rm cos}(F^{x, y}, \mathbf{p}^q)\cdot \sigma\big)},
\end{equation}
where $\sigma$ is the hyper-parameter for softmax function and set as $20$ following \cite{wang2019panet}.

For images with annotated pseudo masks, an auxiliary decoding branch is added after the encoding network to learn the pseudo masks directly. Typical cross-entropy loss for semantic segmentation tasks are utilized to learn from pseudo labeled images, which is denoted as $\mathcal{L}_{pseudo}$. The total loss for training our model is described in Eq. (\ref{loss}).

\textbf{Exponential moving average.}
In our experiments, we find that, with the extra supervision of pseudo masks, the model converges much faster. And the noisy pseudo masks tend to oscillate the performance in later stages. Therefore, we maintain an exponential moving average of model parameters \cite{tarvainen2017mean} to obtain a more stable model for evaluation.

\subsection{Rectifying Support Prototypes}

One challenge of few-shot learning lies in limited annotated samples when adapting to novel classes. To alleviate the problem, we rectify background and foreground prototypes respectively. 

\vspace{0.2cm}
\noindent
\textbf{Global background prototype.}
The typical practice in few-shot segmentation is to extract background prototype from the background regions of current support classes. We assume, however, the characteristics of backgrounds not strongly conjugated with particular foreground classes. In view of this, we propose to incorporate the current support background prototype $\mathbf{p}_{bg}^{cur}$ with a more stable global background prototype $\mathbf{p}_{bg}^{global}$, which is a exponential moving average of all background prototypes learnt during training. Specifically, the global background prototype is updated iteratively during training by:
\vspace{-0.05cm}
\begin{equation}
\vspace{-0.05cm}
    \mathbf{p}_{bg}^{global} \leftarrow m \mathbf{p}_{bg}^{global} + (1-m) \mathbf{p}_{bg}^{cur},
\end{equation}
where $m$ is the momentum coefficient and set as $0.999$ by default for a stable evolution. During training, we keep an additional memory space to store $\mathbf{p}_{bg}^{global}$, and use it for background classification in our FSS episodic training. 

During inference, we keep the same usage of $\mathbf{p}_{bg}^{global}$ for novel classes. We generate the final background prototype for the novel class as follows:
\vspace{-0.05cm}
\begin{equation}
\vspace{-0.05cm}
    \mathbf{p}_{bg}^{final} = w \mathbf{p}_{bg}^{global} + (1-w) \mathbf{p}_{bg}^{cur},
\end{equation}
where $w$ is the fusion weight and set as $0.9$ to respect the stable and informative global one. The global background prototype encodes various scenes in the dataset, and provides good rectification for current backgrounds. We also tried an offline global background prototype, generated by averaging all the background features on the final model, which however performs worse than the online updated one. This could due to inconsistency between training and testing. 
\vspace{0.2cm}

\noindent
\textbf{Rectifying foreground prototype.} 
Inspired by \cite{liu2019prototype}, during inference we utilize the pseudo labeled regions to rectify the foreground prototypes on an image set, such as training set. Compared with \cite{liu2020part} that leverages superpixels, our method based on regions is more stable.

Given a support image $I^s$, we first select top-$N$ relevant images by measuring cosine similarity of the image embeddings. Within this image pool, we then find out $K$ most relevant regions by measuring the cosine similarity between the region embedding $\mathbf{p}^r_i$ and the support foreground prototype. Here we acquire the image and region embedding both by average pooling on the layer3 of ResNet-50/101. Finally, the foreground prototype is rectified by:
\vspace{-0.1cm}
\begin{equation}
\vspace{-0.15cm}
    \mathbf{p}^s \leftarrow (1 - \beta)\mathbf{p}^s + \beta \sum_i \mu_i \mathbf{p}^r_i,
\end{equation}
where $\mathbf{p}^r_i$ is the most relevant region-level prototype in the $i$-th image. $\beta$ is the rectification weight. $\mu_i$ measures the relative similarity between all region-level prototypes and support prototype. It is computed by:
\vspace{-0.15cm}
\begin{equation}
    \mu_i = \frac{{\rm cos}(\mathbf{p}^r_i, \mathbf{p}^s)}{\sum_j {\rm cos}(\mathbf{p}^r_j, \mathbf{p}^s)}.
\end{equation}

\section{Experiments}

\begin{table*}
\caption{Mean IOU of 1-way on PASCAL-$5^i$. The result of PANet with ResNet-50 backbone is obtained from PPNet \cite{liu2020part}. The number of parameters reported in the last column is computed during testing time. The best performance and least parameters are highlighted in bold.}
\renewcommand\arraystretch{0.9}
\small
\centering
\begin{tabular}{l|c|ccccc|ccccc|r}
\specialrule{1.2pt}{2pt}{2pt}
\multicolumn{1}{l|}{\multirow{2}*{Method}} & \multirow{2}*{Backbone} & \multicolumn{5}{c|}{1-shot} & \multicolumn{5}{c|}{5-shot} & \multicolumn{1}{c}{\multirow{2}*{Params}} \\
~ & ~ & fold1 & fold2 & fold3 & fold4 & \textbf{Mean} & fold1 & fold2 & fold3 & fold4 & \textbf{Mean} & \\
\midrule

PGNet \cite{zhang2019pyramid} & \multirow{9}*{ResNet-50} & 56.0 & 66.9 & 50.6 & 50.4 & 56.0 & 54.9 & 67.4 & 51.8 & 53.0 & 56.8 & 32.5 M\\
PANet \cite{wang2019panet} & ~ & 44.0 & 57.5 & 50.8 & 44.0 & 49.1 & 55.3 & 67.2 & 61.3 & 53.2 & 59.3 & 23.5 M\\
CANet \cite{zhang2019canet} & ~ & 52.5 & 65.9 & 51.3 & 51.9 & 55.4 & 55.5 & 67.8 & 51.9 & 53.2 & 57.1 & 36.4 M\\
PPNet \cite{liu2020part} & ~ & 48.6 & 60.6 & 55.7 & 46.5 & 52.8 & 58.9 & 68.3 & 66.8 & 58.0 & 63.0 & 31.5 M\\
PMMs \cite{yang2020prototype} & ~ & 55.2 & 66.9 & 52.6 & 50.7 & 56.3 & 56.3 & 67.3 & 54.5 & 51.0 & 57.3 & 19.6 M\\
PFENet \cite{tian2020prior} & ~ & 61.7 & 69.5 & 55.4 & 56.3 & 60.8 & 63.1 & 70.7 & 55.8 & 57.9 & 61.9 & 34.3 M\\
\textbf{Ours} & ~ & 59.2 & 71.2 & 65.6 & 52.5 & \textbf{62.1} & 63.5 & 71.6 & 71.2 & 58.1 & \textbf{66.1} & \textbf{8.7 M}\\
\textbf{Ours + unlabeled} & ~ & 60.4 & 72.3 & 67.9 & 53.6 & \textbf{63.6} & 64.0 & 72.6 & 71.9 & 58.7 & \textbf{66.8} & \textbf{8.7 M}\\

\specialrule{0pt}{1pt}{1pt}
\hline
\specialrule{0pt}{1pt}{1pt}
\hline
\specialrule{0pt}{1pt}{1pt}

FWB \cite{nguyen2019feature} & \multirow{6}*{ResNet-101} & 51.3 & 64.5 & 56.7 & 52.2 & 56.2 & 54.8 & 67.4 & 62.2 & 55.3 & 59.9 & 43.0 M\\
PPNet \cite{liu2020part} & ~ & 52.7 & 62.8 & 57.4 & 47.7 & 55.2 & 60.3 & 70.0 & 69.4 & 60.7 & 65.1 & 50.5 M\\
DAN \cite{wang2020few} & ~ & 54.7 & 68.6 & 57.8 & 51.6 & 58.2 & 57.9 & 69.0 & 60.1 & 54.9 & 60.5 & -\\
PFENet \cite{tian2020prior} & ~ & 60.5 & 69.4 & 54.4 & 55.9 & 60.1 & 62.8 & 70.4 & 54.9 & 57.6 & 61.4 & 53.4 M\\
\textbf{Ours} & ~ & 60.8 & 71.3 & 61.5 & 56.9 & \textbf{62.6} & 65.8 & 74.9 & 71.4 & 63.1 & \textbf{68.8} & \textbf{27.7 M}\\
\textbf{Ours + unlabeled} & ~ & 61.7 & 72.4 & 63.4 & 57.6 & \textbf{63.8} & 66.2 & 75.4 & 72.0 & 63.4 & \textbf{69.3} & \textbf{27.7 M}\\
\specialrule{1.2pt}{2pt}{2pt}
\end{tabular}
\label{voc_1way_sota}
\vspace{-0.4cm}
\end{table*}

\subsection{Setup}

\textbf{Dataset.}
We evaluate our method extensively on two benchmark datasets, \ie the PASCAL-$5^i$ and COCO-$20^i$. 
The PASCAL-$5^i$ dataset~\cite{shaban2017one} contains 20 categories, which is constructed by PASCAL VOC 2012 \cite{everingham2010pascal} and augmented SBD \cite{BharathICCV2011}. The COCO-$20^i$~\cite{siam2019amp, wang2019panet}, that is a more challenging dataset modified from MS COCO \cite{lin2014microsoft}, consists of $80$ categories.
On both the datasets, we follow the category partition in \cite{wang2019panet}, in which all categories are split into $4$ folds evenly for cross validation. Particularly, three folds are used for training and the remaining one is for evaluation.

\textbf{Network structure.} 
To demonstrate the effectiveness of our method, we utilize the plain network structures, \ie ResNet-50 and ResNet-101, without enhancement designs for evaluation, \eg multi-scale testing.
The last stage is removed for better generalization \cite{yosinski2014transferable} and the last ReLU is removed to measure cosine similarity.
As for the auxiliary mining branch to learn pseudo masks, we simply adopt a lightweight segmentation head which is constructed with three convolution layers, where each convolution is followed by a batch normalization and ReLU except the last one. For a fair comparison with previous methods, we use ImageNet pre-trained ResNet parameters for initialization.

\textbf{Implementation details.}
To re-annotate training images or annotate unlabeled images, we group $5$ clusters on PASCAL-$5^i$ and $15$ clusters on COCO-$20^i$, respectively, by $K$-Means algorithm according to the statistics of the average object number per image on these datasets. Given the groundtruth masks, we train our model following the setting below. In particular, on PASCAL-$5^i$ and COCO-$20^i$, we construct each mini-batch with $4$ support-query pairs and $32$ extra training images supervised by our pseudo masks. Limited by the GPU memory, the number of extra training images is set to $16$ on ResNet-101 in the 5-shot setting. We use the SGD optimizer for training, where the learning rate is initialized by $1e\minus 3$ and decays by $10$ times every $2000$ iterations, and the momentum is $0.9$. A total of $6000$ iterations are optimized. Note that, our training images together with the masks are all cropped to $(473, 473)$ and augmented by random horizontal flipping. The images for learning pseudo masks are strongly augmented following \cite{chen2020simple}. During the evaluation, we follow \cite{tian2020prior} to sample $1000$ and $4000$ support-query pairs on PASCAL-$5^i$ and COCO-$20^i$ respectively, and we run the test with 5 different random seeds and provide their average mean IOU as a stable result. The testing images are evaluated on their original resolution. 

\textbf{Baseline and metrics.}
Since our method is metric learning based, we adopt the same baseline method in PANet \cite{wang2019panet} as our baseline model, which is a metric learning framework consisting of only an encoder. Following \cite{shaban2017one, wang2019panet, liu2020part, liu2020crnet},  we adopt mean Intersection-over-Union (mIOU) for performance evaluation.

\begin{table*}
\caption{Mean IOU of 1-way on COCO-$20^i$. The result of PANet with ResNet-50 backbone is obtained from PPNet \cite{liu2020part}.}
\renewcommand\arraystretch{0.9}
\small
\centering
\begin{tabular}{l|c|ccccc|ccccc|r}
\specialrule{1.2pt}{2pt}{2pt}
\multicolumn{1}{l|}{\multirow{2}*{Method}} & \multirow{2}*{Backbone} & \multicolumn{5}{c|}{1-shot} & \multicolumn{5}{c|}{5-shot} & \multirow{2}*{Params}\\
~ & ~ & fold1 & fold2 & fold3 & fold4 & \textbf{Mean} & fold1 & fold2 & fold3 & fold4 & \textbf{Mean} & ~\\
\midrule
PANet \cite{wang2019panet} & \multirow{4}*{ResNet-50} & 31.5 & 22.6 & 21.5 & 16.2 & 23.0 & 45.9 & 29.2 & 30.6 & 29.6 & 33.8 & 23.5 M\\
PPNet \cite{liu2020part} & ~ & 36.5 & 26.5 & 26.0 & 19.7 & 27.2 & 48.9 & 31.4 & 36.0 & 30.6 & 36.7 & 31.5 M\\
\textbf{Ours} & ~ & 46.8 & 35.3 & 26.2 & 27.1 & \textbf{33.9} & 54.1 & 41.2 & 34.1 & 33.1 & \textbf{40.6} & \textbf{8.7 M}\\
\textbf{Ours + unlabeled} & ~ & 48.0 & 36.6 & 27.4 & 28.2 & \textbf{35.1} & 54.9 & 42.1 & 34.9 & 33.6 & \textbf{41.4} & \textbf{8.7 M}\\

\specialrule{0pt}{1pt}{1pt}
\hline
\specialrule{0pt}{1pt}{1pt}
\hline
\specialrule{0pt}{1pt}{1pt}

PMMs \cite{yang2020prototype} & \multirow{4}*{ResNet-101} & 29.5 & 36.8 & 28.9 & 27.0 & 30.6 & 33.8 & 42.0 & 33.0 & 33.3 & 35.5 & 38.6 M\\
PFENet \cite{tian2020prior} & ~ & 34.3 & 33.0 & 32.3 & 30.1 & 32.4 & 38.5 & 38.6 & 38.2 & 34.3 & 37.4 & 53.4 M\\
\textbf{Ours} & ~ & 50.2 & 37.8 & 27.1 & 30.4 & \textbf{36.4} & 57.0 & 46.2 & 37.3 & 37.2 & \textbf{44.4} & \textbf{27.7 M}\\
\textbf{Ours + unlabeled} & ~ &51.1 & 38.7 & 28.5 & 31.6 & \textbf{37.5} & 57.8 & 47.1 & 37.8 & 37.6 & \textbf{45.1} & \textbf{27.7 M}\\
\specialrule{1.2pt}{2pt}{2pt}
\end{tabular}
\label{coco_1way_sota}
\vspace{-0.4cm}
\end{table*}

\begin{table}
\caption{Ablation studies on the effect of different components. \textbf{FG}: Rectify foreground prototype with most relevant regions. \textbf{BG}: Rectify background prototype by incorporating global background prototype. \textbf{Mine}: Mine latent classes from training images.}
\renewcommand\arraystretch{0.9}
\small
\centering
\begin{tabular}{cccccccc}

\specialrule{1.2pt}{2pt}{2pt}

FG & BG & Mine & fold1 & fold2 & fold3 & fold4 & \textbf{Mean}\\

\specialrule{0pt}{1pt}{1pt}
\hline
\specialrule{0pt}{1pt}{1pt}

 & & & 56.4 & 66.4 & 60.6 & 47.7 & 57.8 \\
\cmark &  &  & 58.2 & 68.6 & 60.7 & 48.2 & 58.9\\
 & \cmark &  & 55.7 & 68.3 & 63.1 & 49.9 & 59.3\\
 & & \cmark & 58.1 & 68.9 & 63.8 & 48.3 & 59.8\\

\specialrule{0pt}{0.5pt}{0.5pt}
\hline
\specialrule{0pt}{1.5pt}{1.5pt}

\cmark & \cmark &  & 57.6 & 70.3 & 63.2 & 50.6 & 60.4\\
\cmark & \cmark & \cmark & \textbf{59.2} & \textbf{71.2} & \textbf{65.6} & \textbf{52.5} & \textbf{62.1}\\

\specialrule{1.2pt}{2pt}{2pt}
\end{tabular}

\label{ablation_module}
\vspace{-0.3cm}
\end{table}

\subsection{Comparison with State-of-the-Arts}
We evaluate the effectiveness of our method on two benchmark datasets \cite{shaban2017one,everingham2010pascal, BharathICCV2011}. 
In particular, we conduct extensive experiments with the widely-used encoding networks, \ie ResNet-50 and ResNet-101, on various few-shot segmentation settings, which includes $1$-shot and $5$-shot on $1$-way. Here, K-shot N-way indicates k samples for each category of the N categories. Extensive experiments show our superiority to the previous methods in all cases. 

\begin{figure}
\vspace{-0.2cm}
    \centering
    \includegraphics[width=0.4\textwidth]{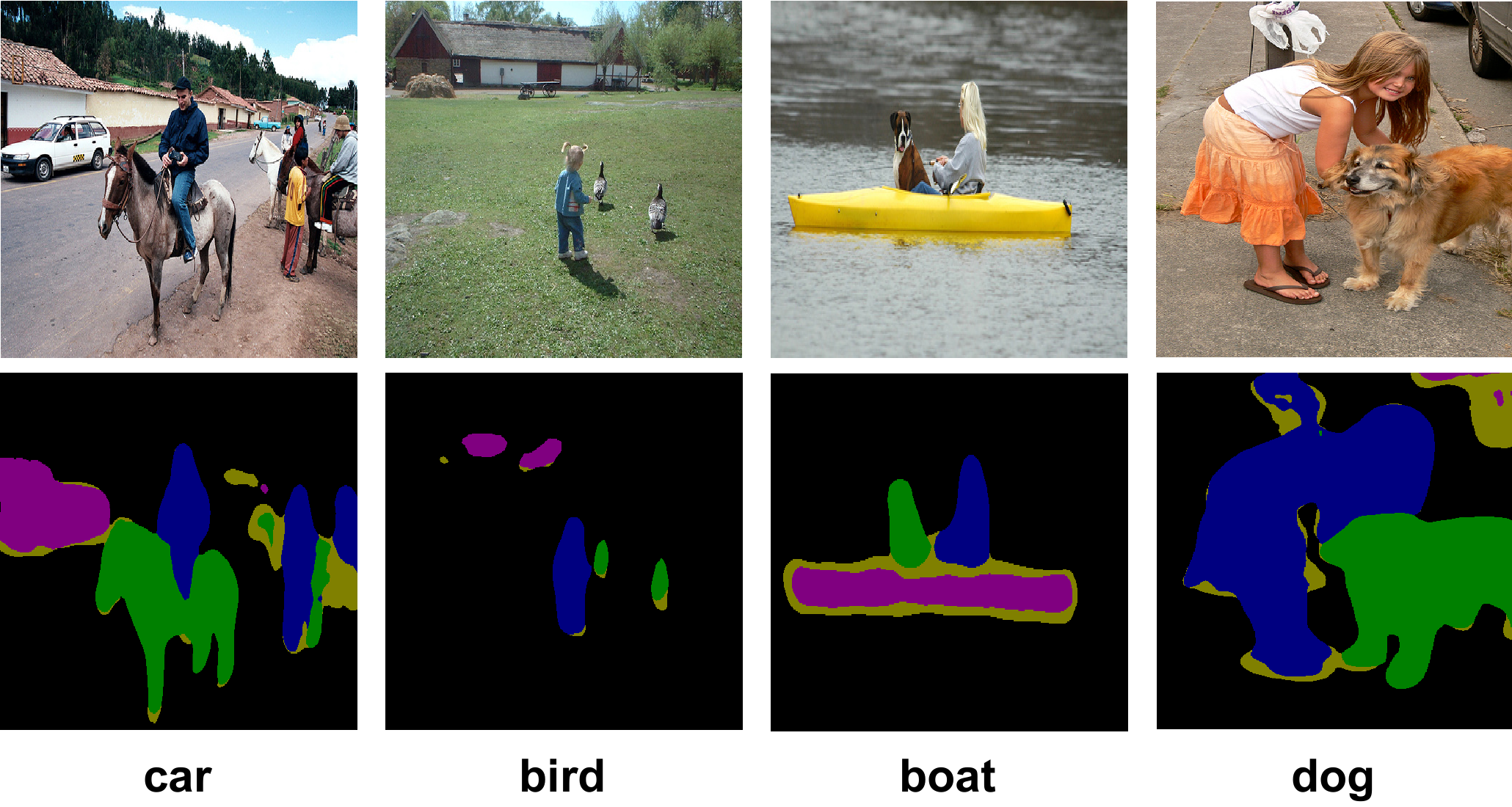}
    \caption{Visualization of pseudo masks on PASCAL-$5^i$. Note that, the colors in pseudo masks only stand for abstract semantic categories of sub-clusters rather than any concrete semantics. The last line illustrates the mined latent novel classes.}
    \label{pseudo}
    \vspace{-0.3cm}
\end{figure}

\begin{figure}
\vspace{-0.2cm}
    \centering
    \includegraphics[width=0.4\textwidth]{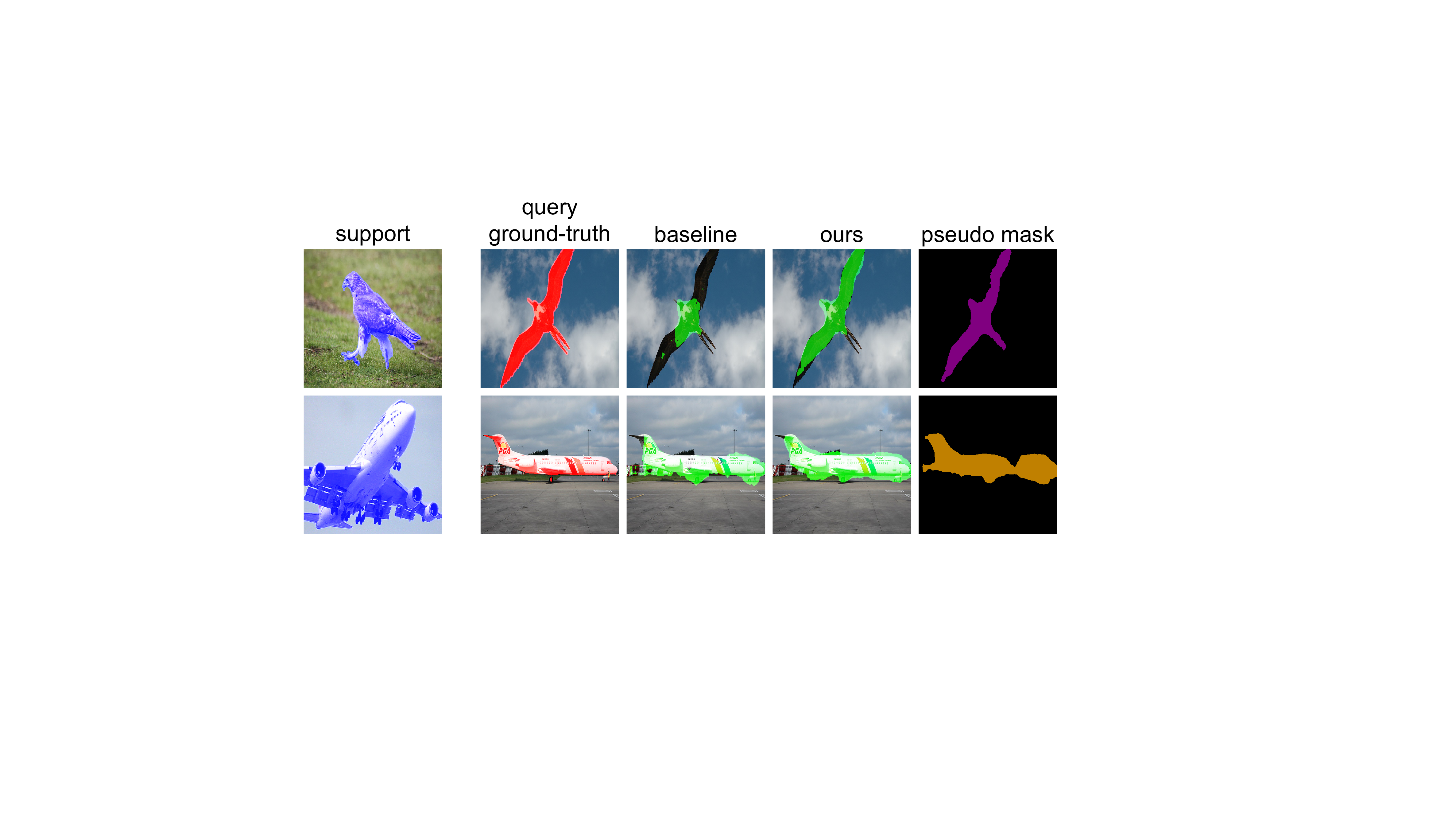}
    \caption{Visualization of 1-way 1-shot setting on PASCAL-$5^i$.}
    \label{prediction}
\vspace{-0.4cm}
\end{figure}

\textbf{PASCAL-$\mathbf{5^i}$.}
From Table \ref{voc_1way_sota}, we can see that, on both the ResNet-50 and ResNet-101, our method outperforms previous state-of-the-art by a large margin in both 1-shot and 5-shot setting with the fewest parameters among all existing methods. Specifically, in the 1-shot setting, our method surpasses the state-of-the-art by $1.3\%$ and $2.5\%$ with ResNet-50 and ResNet-101 respectively. And our method performs significantly better than other methods by $3.1\%$ and $3.7\%$ with the two backbones respectively in the $5$-shot setting, showing its effectiveness in multi-shot cases. With all these improvements, our method even takes $74\%$ fewer parameters than the previous state-of-the-art. Moreover, our method can be further boosted with extra unlabeled data, which is the remaining images without any base classes from original training set. The effect of unlabeled data is discussed in detail in Section \ref{ablation_sec}. The visualization of pseudo masks and predictions is shown in Figure \ref{pseudo} and Figure \ref{prediction}. The annotated pseudo masks can mine the latent novel classes from the backgrounds as expected, which further proved the effectiveness of our pseudo labeling process.

\textbf{COCO-$\mathbf{20^i}$.} 
The COCO-$20^i$ dataset is a very challenging dataset that usually contains many objects in a realistic scene image. On this dataset, we outperform previous best results by a large margin of $6.7\%$ and $4.0\%$ on the challenging $1$-shot setting with ResNet-50 and ResNet-101 respectively, as shown in Table \ref{coco_1way_sota}. In addition, we gain an impressive improvement of $3.9\%$ and $7.0\%$ in the $5$-shot setting with our two backbones respectively, which proves the superiority of our method in such challenging scenarios.

\subsection{Ablation Study} \label{ablation_sec}

We conduct extensive ablation studies with ResNet-50 in the $1$-way $1$-shot setting on PASCAL-$5^i$.

\textbf{Effectiveness of different components.}
Our method contains three components, namely mining latent novel classes, rectifying the background prototype and rectifying the foreground prototype. We validate the effectiveness of each component in Table \ref{ablation_module}. It is shown that the mining of latent classes plays the most important role in the performance improvement, and meantime the rectification technique for support prototypes is indispensable. With all the three components, our method achieves the best performance and surpasses the state-of-the-art.

\begin{table}
\caption{Comparisons with advanced semi-supervised and self-supervised methods. \textbf{Training Set}: the same training set in FSS \textbf{(no extra data are introduced)}. \textbf{Remaining}: the remaining raw training images without any base classes.}
\renewcommand\arraystretch{0.9}
\small
\centering
\begin{tabular}{cccccc}

\specialrule{1.2pt}{2pt}{2pt}

Method & fold1 & fold2 & fold3 & fold4 & \textbf{Mean}\\

\specialrule{0pt}{1pt}{1pt}
\hline
\specialrule{0pt}{1pt}{1pt}

\multicolumn{6}{c}{Unlabeled Source: Training Set}\\

\specialrule{0pt}{1pt}{1pt}
\hline
\specialrule{0pt}{1pt}{1pt}

CCT \cite{ouali2020semi} & 57.5 & 68.0 & 59.6 & 49.1 & 58.6\\
SimCLR \cite{chen2020simple} & 56.5 & 63.4 & 59.8 & 47.9 & 56.9\\
Ours & 58.1 & 68.9 & 63.8 & 48.3 & \textbf{59.8}\\

\specialrule{0pt}{1pt}{1pt}
\hline
\specialrule{0pt}{1pt}{1pt}

\multicolumn{6}{c}{Unlabeled Source: Training Set + Remaining}\\

\specialrule{0pt}{1pt}{1pt}
\hline
\specialrule{0pt}{1pt}{1pt}

CCT \cite{ouali2020semi} & 58.1 & 68.8 & 59.1 & 49.4 & 58.9\\
SimCLR \cite{chen2020simple} & 56.5 & 64.1 & 61.2 & 48.3 & 57.5\\
Ours & 59.2 & 70.2 & 65.7 & 49.8 & \textbf{61.2}\\

\specialrule{1.2pt}{2pt}{2pt}
\end{tabular}

\label{semi}
\end{table}

\textbf{Effectiveness in leveraging the unlabeled.}
As an extension of our work, we introduce the unlabeled data to our mining module. Based on the plain baseline, we compare our mining methods with advanced self-supervised and semi-supervised methods. 
Here we adopt SimCLR \cite{chen2020simple} and CCT \cite{ouali2020semi} as the feature mining method, and leverage their trained backbone on base classes as the feature extractor for support-query pairs. Later the same inference strategy as ours are conducted.
For the self-supervised learning method SimCLR \cite{chen2020simple}, we apply the contrastive learning on both the training data and unlabeled data. And for the semi-supervised learning method CCT \cite{ouali2020semi}, it is trained on the annotated base classes as well as the background and exploits the unlabeled within the pre-defined classes. In Table \ref{semi}, we first examine the scenario where no extra data is introduced, which means that the training images are both treated as the labeled and unlabeled. And then we leverage the remaining images without their annotations in original training set for semantic segmentation task. As shown in Table \ref{semi}, in both scenarios, our method are consistently more effective in exploiting the training data and extra unlabeled data. The worse performance of SimCLR further shows that the invariance constraint on different crops of an image is not appropriate to the dense prediction task.

\begin{table}
\caption{Ablation studies on different sources of unlabeled data. \textbf{None}: without rectifying the foreground class and mining, only rectifying global background prototype. \textbf{Trainset}: the same training set in FSS \textbf{(no extra data are introduced)}. \textbf{Trainset + Remain}: the same training set in FSS and the remaining raw training images without any base classes. \textbf{IN}: ImageNet. And for efficient training, we uniformly sample a subset of 10 images per class.}
\renewcommand\arraystretch{0.9}
\small
\centering
\begin{tabular}{cccccc}

\specialrule{1.2pt}{2pt}{2pt}

Unlabeled Source & fold1 & fold2 & fold3 & fold4 & \textbf{Mean}\\

\specialrule{0pt}{1pt}{1pt}
\hline
\specialrule{0pt}{1pt}{1pt}

None & 55.7 & 68.3 & 63.1 & 49.9 & 59.3\\
ImageNet & 59.4 & 70.6 & 64.8 & 52.7 & 61.9\\
Trainset & 59.2 & 71.2 & 65.6 & 52.5 & 62.1\\

\specialrule{0pt}{1pt}{1pt}
\hline
\specialrule{0pt}{1pt}{1pt}

Trainset + IN & 59.8 & 71.8 & 66.1 & 53.3 & 62.8\\
Trainset + Remain & 60.4 & 72.3 & 67.9 & 53.6 & 63.6\\

\specialrule{1.2pt}{2pt}{2pt}
\end{tabular}

\label{source}
\end{table}

\begin{table}
\vspace{-0.1cm}
\caption{Ablation studies on the effect of exponential moving average (EMA) of model parameters \cite{tarvainen2017mean}. \textbf{Full}: the overall method.}
\renewcommand\arraystretch{0.9}
\small
\centering
\begin{tabular}{cccccc}

\specialrule{1.2pt}{2pt}{2pt}

Method & fold1 & fold2 & fold3 & fold4 & \textbf{Mean}\\

\specialrule{0pt}{1pt}{1pt}
\hline
\specialrule{0pt}{1pt}{1pt}

Baseline & 56.4 & 66.4 & 60.6 & 47.7 & 57.8\\
Baseline \textit{w/} EMA & 56.0 & 66.2 & 61.9 & 47.6 & 57.9\\

\specialrule{0pt}{1pt}{1pt}
\hline
\specialrule{0pt}{1pt}{1pt}

FG + BG & 57.6 & 70.3 & 63.2 & 50.6 & 60.4\\
FG + BG \textit{w/} EMA & 57.5 & 70.1 & 63.9 & 50.4 & 60.5\\

\specialrule{0pt}{1pt}{1pt}
\hline
\specialrule{0pt}{1pt}{1pt}

Full \textit{w/o} EMA & 58.5 & 70.8 & 64.2 & 52.1 & 61.4\\
Full \textit{w/} EMA & 59.2 & 71.2 & 65.6 & 52.5 & 62.1\\

\specialrule{1.2pt}{2pt}{2pt}
\end{tabular}
\label{ema}
\end{table}

\textbf{Different sources of unlabeled data.}
To examine the performance under different sources of unlabeled data, we compare the effects of different data sources in Table \ref{source}. 
Considering our encoding network is initialized with the pre-trained weight on ImageNet, the $2.6\%$ performance gain proves our re-use of the data is effective. In addition, by treating labeled training images as our unlabeled images, we could boost the performance of our method by $2.8\%$. Moreover, by combining different sources of data, our method can further be improved. That proves the effectiveness of our method in mining latent novel classes again.

\textbf{Effect of the EMA.}
Considering the fast convergence of training process when supervised by pseudo masks and the oscillation caused by noisy labels, we use the exponential moving average (EMA) technique \cite{tarvainen2017mean} to obtain a stable model for evaluation. Therefore, we add EMA to our method of different versions in Table \ref{ema} and find that the mining module benefits much more from the EMA than our baseline models. It further proves our observation is correct and the corresponding solution is effective.

\textbf{Efficiency of our method.}
Our method surpasses the state-of-the-art by a large margin at the cost of much fewer parameters and much faster inference speed. Specifically, in Table \ref{fps}, our model takes only $8.7$M parameters compared with the $34.3$M of PFENet \cite{tian2020prior}. Besides, the inference speed of our method is 1.8x and 2.5x faster than PFENet in 1-shot and 5-shot setting respectively.

\textbf{Hyper-parameters.}
Except the widely adopted hyper-parameters of previous methods, such as the $\sigma$ in the softmax function, the rest of the hyper-parameters are examined on the left classes in the training set. The rectification weight $\beta$ is set as $0.2$ and the number $N$ of selected relevant images is set as $4$ since we find that the larger $N$ will bring more noise and increase the inference time. We show the ablations on the most important hyper-parameter $K$ in the K-Means algorithm in Table \ref{K}.

\begin{table}
\caption{Frames (number of episodes) per second and number of parameters.}
\renewcommand\arraystretch{0.9}
\small
\centering
\begin{tabular}{cccccr}

\specialrule{1.2pt}{2pt}{2pt}

\multirow{2}*{Method} & \multicolumn{2}{c}{1-shot} & \multicolumn{2}{c}{5-shot} & \multirow{2}*{Params}\\
~ & FPS & mIOU & FPS & mIOU & \\

\specialrule{0pt}{1pt}{1pt}
\hline
\specialrule{0pt}{1pt}{1pt}
PMMs \cite{yang2020prototype} & 18.2 & 56.3 & 9.4 & 57.3 & 19.6 M\\
PFENet \cite{tian2020prior} & 15.7 & 60.8 & 5.1 & 61.9 & 34.3 M\\
Ours & \textbf{27.8} & \textbf{62.1} & \textbf{12.8} & \textbf{66.1} & \textbf{8.7 M}\\

\specialrule{1.2pt}{2pt}{2pt}
\end{tabular}

\label{fps}
\end{table}

\begin{table}
\vspace{-0.2cm}
\caption{Ablation studies on the $K$ in K-Means.}
\renewcommand\arraystretch{0.9}
\small
\centering
\begin{tabular}{cccccc}

\specialrule{1.2pt}{2pt}{2pt}

K & fold1 & fold2 & fold3 & fold4 & \textbf{Mean}\\

\specialrule{0pt}{1pt}{1pt}
\hline
\specialrule{0pt}{1pt}{1pt}

1 & 57.6 & 69.3 & 64.3 & 50.2 & 60.4\\
3 & 58.7 & 70.2 & 65.1 & 51.1 & 61.3\\
5 & \textbf{59.2} & \textbf{71.2} & \textbf{65.6} & \textbf{52.5} & \textbf{62.1}\\
7 & 58.9 & 70.4 & 64.7 & 51.5 & 61.4\\
9 & 58.1 & 69.6 & 64.1 & 50.6 & 60.6\\

\specialrule{1.2pt}{2pt}{2pt}
\end{tabular}

\label{K}
\end{table}

\section{Conclusion}
In this work, we address few-shot segmentation from a novel perspective via mining latent classes from the backgrounds and propose a novel framework to learn meta knowledge as well as mine good embedding from both the groundtruth masks and our pseudo masks. Above this, we propose a novel rectification technique for support prototypes. Extensive experiments are conducted on two FSS benchmarks and without bells and whistles, our method can outperform previous methods by a large margin. Moreover, through ablation studies and the comparison with advanced self-supervised and semi-supervised learning techniques, our method can better exploit the knowledge in the training data via mining latent novel classes behind them.

{\small
\bibliographystyle{ieee_fullname}
\bibliography{egbib}
}

\clearpage

\end{document}